\documentclass[11pt,a4paper]{article}
\usepackage[hyperref]{emnlp2020}
\usepackage{times}
\usepackage{latexsym}

\usepackage{microtype}
\usepackage{booktabs}
\usepackage{multirow}
\usepackage{algorithm,algcompatible,amsmath}
\usepackage{stackengine,scalerel,mathtools}
\usepackage{url}
\usepackage{amssymb}
\usepackage{graphicx}
\usepackage{mathtools}
\usepackage{xcolor}
\usepackage{enumitem}
\usepackage{marvosym}
\usepackage{textcomp}
\usepackage{array}

\usepackage{microtype}

\aclfinalcopy 

\title{Lightweight, Dynamic Graph Convolutional Networks \\
for AMR-to-Text Generation}

\author{Yan Zhang\thanks{$^{*}$ Equally Contributed. Work done while Yan Zhang was an intern at DAMO Academy, Alibaba Group and Zhijiang Guo was at the University of Edinburgh.}$^{*1,2}$~~~ Zhijiang Guo$^{*1}$~~~ Zhiyang Teng$^{3}$~~~Wei Lu$^{1}$
\\ \textbf{Shay B. Cohen}$^{4}$~~~\textbf{Zuozhu Liu}$^{5}$~~~ \textbf{Lidong Bing}\thanks{~~Corresponding author.}$^{~2}$\\
  $^{1}$Singapore University of Technology and Design \\ $^{2}$DAMO Academy, Alibaba Group,  $^{3}$Westlake University  \\
  $^{4}$University of Edinburgh, $^{5}$ZJU-UIUC Institute\\
   \texttt{\{yan\_zhang,zhijiang\_guo\}@mymail.sutd.edu.sg, luwei@sutd.edu.sg}\\
  \texttt{tengzhiyang@westlake.edu.cn, scohen@inf.ed.ac.uk}\\
  \texttt{zuozhuliu@intl.zju.edu.cn, l.bing@alibaba-inc.com}}

\date{}

\begin{document}
\maketitle

\begin{abstract}
	
	AMR-to-text generation is used to transduce Abstract Meaning Representation structures (AMR) into text. A key challenge in this task is to efficiently learn effective graph representations. Previously, Graph Convolution Networks (GCNs) were used to encode input AMRs, however, vanilla GCNs are not able to capture non-local information and additionally, they follow a local (first-order) information aggregation scheme. To account for these issues, larger and deeper GCN models are required to capture more complex interactions. In this paper, we introduce a dynamic fusion mechanism, proposing Lightweight Dynamic Graph Convolutional Networks (LDGCNs) that capture richer non-local interactions by synthesizing higher order information from the input graphs. We further develop two novel parameter saving strategies based on the group graph convolutions and weight tied convolutions to reduce memory usage and model complexity. With the help of these strategies, we are able to train a model with fewer parameters while maintaining the model capacity. Experiments demonstrate that LDGCNs outperform state-of-the-art models on two benchmark datasets for AMR-to-text generation with significantly fewer parameters.
\end{abstract}

\section{Introduction}
\label{sec:1}

\begin{figure}[!t]
	\centering
	\includegraphics[scale=0.42]{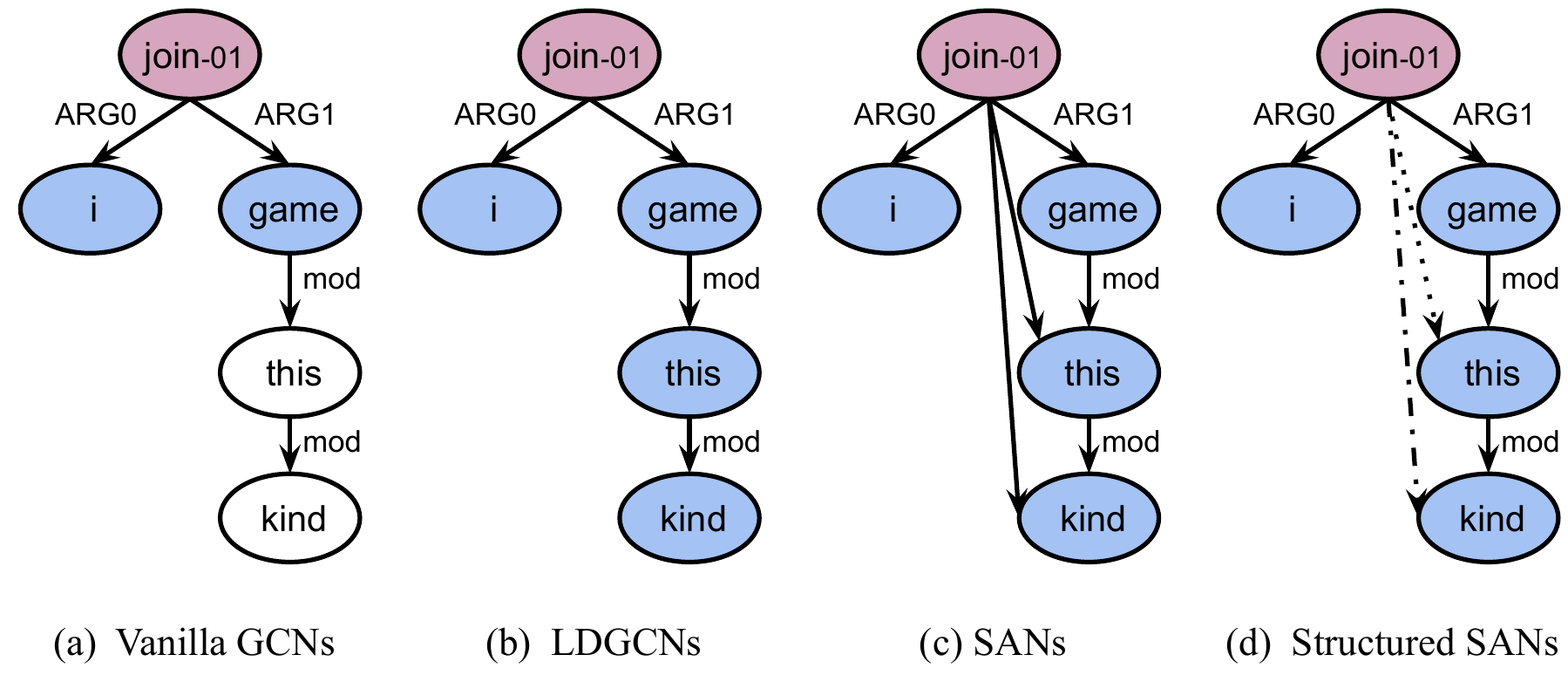}
	\vspace{-6mm}
	\caption{\protect The concept (\textit{join-01}) in vanilla GCNs is that it only captures information from its immediate neighbors (first-order), while in LDGCNs it can integrate information from neighbors of different order  (e.g., second-order and third-order). In SANs, the node collects information from all other nodes, while in structured SANs it is aware of its connected nodes in the original graph.}
	\vspace{-6mm}
	\label{fig:amr}
\end{figure}

Graph structures play a pivotal role in NLP because they are able to capture particularly rich structural information. For example, Figure~\ref{fig:amr} shows a directed, labeled Abstract Meaning Representation (AMR; \citealt{Banarescu2013AbstractMR}) graph, where each node denotes a semantic concept and each edge denotes a relation between such concepts. Within the realm of work on AMR, we focus in this paper on the problem of AMR-to-text generation, i.e. transducing AMR graphs into text that conveys the information in the AMR structure. A key challenge in this task is to efficiently learn useful representations of the AMR graphs. Early efforts \citep{Pourdamghani2016GeneratingEF, Konstas2017NeuralAS} neglect a significant part of the structural information in the input graph by linearizing it. Recently, Graph Neural Networks (GNNs) have been explored to better encode structural information for this task \citep{Beck2018GraphtoSequenceLU,Song2018AGM,Damonte2019StructuralNE,Ribeiro2019EnhancingAG}. 


One type of such GNNs is Graph Convolutional Networks (GCNs; \citealt{Kipf2016SemiSupervisedCW}).  GCNs follow a local information aggregation scheme, iteratively updating the representations of nodes based on their immediate (first-order) neighbors.  Intuitively, stacking more convolutional layers in GCNs helps capture more complex interactions \citep{Xu2018RepresentationLO, guo2019densely}.  However, prior efforts \citep{Zhu2019ModelingGS, Cai2019GraphTF, Wang2020AMRToTextGW} have shown that the locality property of existing GCNs precludes efficient non-local information propagation. \citet{AbuElHaija2019MixHopHG} further proved that vanilla GCNs are unable to capture feature differences among neighbors from different orders no matter how many layers are stacked. Therefore, Self-Attention Networks (SANs; \citealt{Vaswani2017AttentionIA}) have been explored as an alternative to capture global dependencies. As shown in Figure~\ref{fig:amr} (c), SANs associate each node with other nodes such that we model interactions between any two nodes in the graph. Still, this approach ignores the structure of the original graph. \citet{Zhu2019ModelingGS} and \citet{Cai2019GraphTF} propose structured SANs that incorporate additional neural components to encode the structural information of the input graph. 

Convolutional operations, however, are more computationally efficient than self-attention operations because the computation of attention weights scales quadratically while convolutions scale linearly with respect to the input length \citep{wu2019pay}. Therefore, it is worthwhile to explore the possibility of models based on graph convolutions. One potential approach that has been considered is to incorporate information from higher order neighbors, which helps to facilitate non-local information aggregation for node classification \citep{abu2018n, AbuElHaija2019MixHopHG, Morris2019WeisfeilerAL}. However, simple concatenation of different order representations may not be able to model complex interactions in semantics for text generation \citep{luan2019break}.

We propose to better integrate high-order information, by introducing a novel dynamic fusion mechanism and propose the Lightweight, Dynamic Graph Convolutional Networks (LDGCNs). As shown in Figure \ref{tab:amr} (b), nodes in the LDGCN model are able to integrate information from first to third-order neighbors. With the help of the dynamic mechanism, LDGCNs can effectively synthesize information from different orders to model complex interactions in the AMR graph for text generation. Also, LDGCNs require no additional computational overhead, in contrast to vanilla GCN models. We further develop two novel weight sharing strategies based on the group graph convolutions and weight tied convolutions. These strategies allow the LDGCN model to reduce memory usage and model complexity.

Experiments on AMR-to-text generation show that LDGCNs outperform best reported GCNs and SANs trained on LDC2015E86 and LDC2017T10 with significantly fewer parameters. On the large-scale semi-supervised setting, our model is also consistently better than others, showing the effectiveness of the model on a large training set. We release our code and pretrained models at \url{https://github.com/yanzhang92/LDGCNs}.\footnote{Our implementation is based on  MXNET \citep{Chen2015MXNetAF} and the Sockeye toolkit \citep{hieber2017sockeye}.}

\section{Background}
\label{sec:2}
\paragraph{Graph Convolutional Networks}
Our LDGCN model is closely related to GCNs \citep{Kipf2016SemiSupervisedCW} which restrict filters to operate on a first-order neighborhood. Given an AMR graph $\mathcal{G}$ with $n$ concepts (nodes),  GCNs associate each concept $v$ with a feature vector $\mathbf{h}_v$ $\in$ $\mathbb{R}^{d}$, where $d$ is the feature dimension. $\mathcal{G}$ can be represented by concatenating features of all the concepts, i.e., $\mathbf{H}$=$[\mathbf{h}_{v_1},...,\mathbf{h}_{v_n}]$. Graph convolutions at $l$-th layer can be defined as:
\begin{equation}
\begin{aligned}
\vspace{-1mm}
\mathbf{H}_{l+1} = \phi (\mathbf{A} \mathbf{H}_{l} \mathbf{W}_l +  \mathbf{b}_l ),
\label{eq:graphconv}
\vspace{-1mm}
\end{aligned}
\end{equation}
where $\mathbf{H}_{l}$ is hidden representations of the $l$-th layer. $\mathbf{W}_l$ and $\mathbf{b}_l$ are trainable model parameters for the $l$-th layer,  $\phi$ is an activation function. $\mathbf{A}$ is the adjacency matrix, $\mathbf{A}_{uv}$=1 if there exists a relation (edge) that goes from concept $u$ to concept $v$.

\paragraph{Self-Attention Networks}
Unlike GCNs, SANs \citep{Vaswani2017AttentionIA} capture global interactions by connecting each concept to all other concepts. Intuitively, the attention matrix can be treated as the adjacency matrix of a fully-connected graph. Formally, SANs take a sequence of representations of $n$ nodes $\mathbf{H}$=$[\mathbf{h}_{v_1},...,\mathbf{h}_{v_n}]$ as the input.
Attention score $\mathbf{A}_{uv}$ between the concept pair ($u$,$v$) is:
\begin{equation}
\begin{aligned}
\vspace{-1mm}
\mathbf{A}_{uv} &= f(\mathbf{h}_{u}, \mathbf{h}_{v}) \\
&= \mathrm{softmax}(\frac{(\mathbf{h}_{u}\mathbf{W}^Q) (\mathbf{h}_{v}\textbf{W}^K)^T}{\sqrt{d}})
\label{eq:multihead}
\vspace{-1mm}
\end{aligned}
\end{equation}
where $\textbf{W}^Q$ and $\textbf{W}^K$ are projection parameters. The adjacency matrix $\mathbf{A}$ in GCNs is given by the input AMR graph, while in SANs $\mathbf{A}$ is computed based on $\mathbf{H}$, which neglects the structural information of the input AMR. The number of operations required by graph convolutions scales is found linearly in the input length, whereas they are quadratic for SANs.

\paragraph{Structured SANs}
\citet{Zhu2019ModelingGS} and \citet{Cai2019GraphTF} extend SAN s
by incorporating the relation $\mathbf{r}_{uv}$ between node $u$ and node $v$ in the original graph such that the model is aware of the input structure when computing attention scores:
\begin{equation}
\begin{aligned}
\vspace{-1mm}
\mathbf{A}_{uv} = g(\mathbf{h}_{u}, \mathbf{h}_{v}, \mathbf{r}_{uv})
\label{eq:relation}
\vspace{-1mm}
\end{aligned}
\end{equation}
where $\mathbf{r}_{uv}$ is obtained based on the shortest relation path between the concept pair $(u, v)$ in the graph. For example, the shortest relation path between (\textit{join-01}, \textit{this}) in Figure \ref{tab:amr} (d) is [ARG1, mod]. Formally, the path between concept $u$ and $v$ is represented as $s_{uv}$=$[e(u, k_{1}), e(k_{1}, k_{2}), . . . , e(k_{m}, v)]$, where $e$ indicates the relation label between two concepts and $k_{1:m}$ are the relay nodes.
We have    $\mathbf{r}_{uv} = f(s_{uv})$
where $f$ is a sequence encoder, and this can be performed with gated recurrent units (GRUs) or convolutional neural networks (CNNs).

\begin{figure}
	\centering
	\includegraphics[scale=0.44]{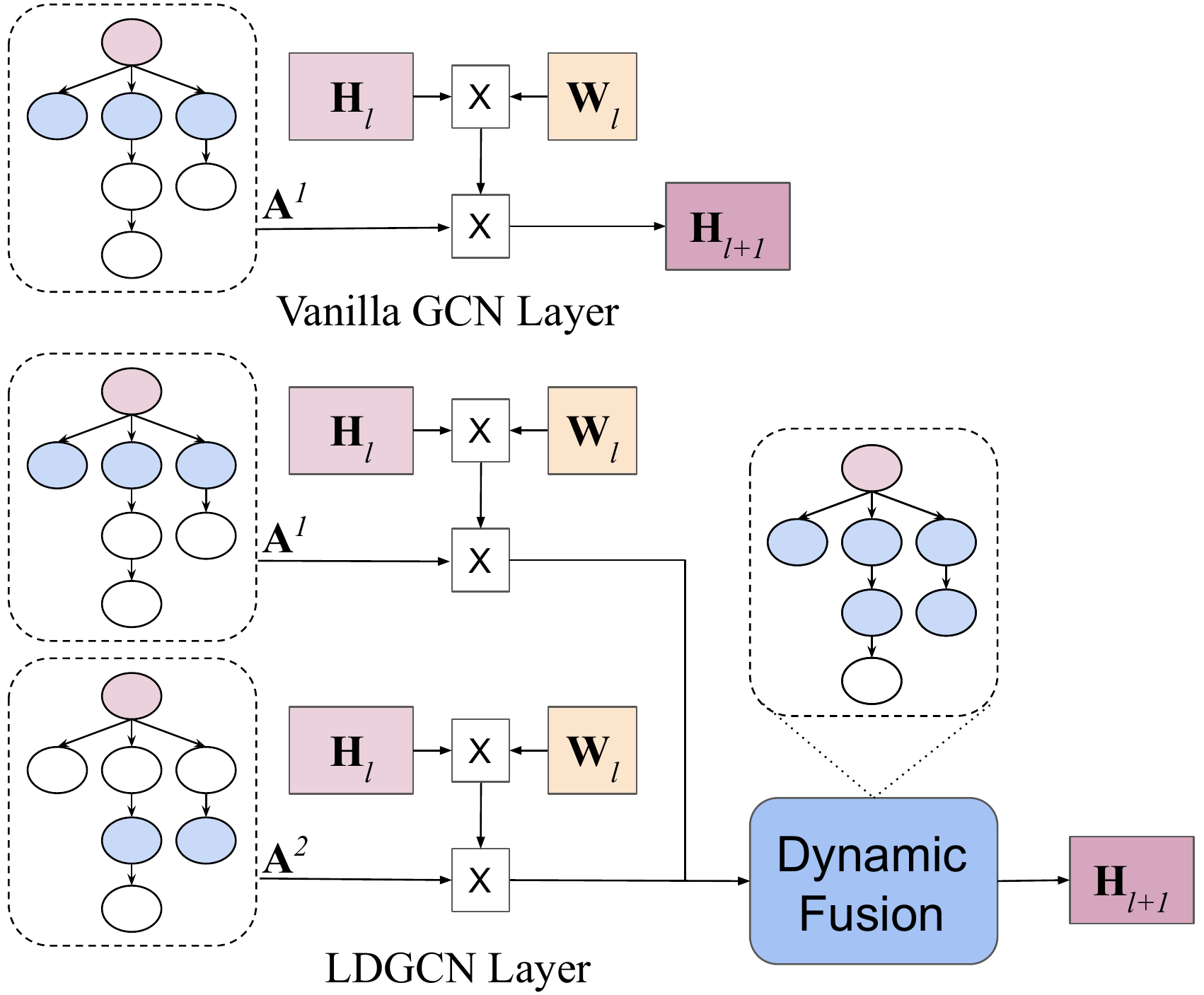}
	\vspace{-1mm}
	\caption{Comparison between vanilla GCNs and LDGCNs. $\mathbf{H}_{l}$ denotes the representation of $l$-th layer. $\mathbf{W}_{l}$ denotes the trainable weights and $\times$ denotes matrix multiplication. Vanilla GCNs take the 1st-order adjacency matrix $\mathbf{A}^{1}$ as the input, which only captures information from one-hop neighbors. LDGCNs take $k$ number of $k$-order adjacency matrix $\mathbf{A}^{k}$ as inputs, $\mathbf{W}_{l}$ is shared for all $\mathbf{A}^{k}$. $k$ is set to  2 here for simplification. A dynamic fusion mechanism is applied to integrate the information from 1- to $k$-hop neighbors.} 
	\vspace{-4mm}
	\label{fig:layercomp}
\end{figure}

\section{Dynamic Fusion Mechanism}
\label{sec:4}
As discussed in Section \ref{sec:2}, GCNs are generally more computationally efficient than structured SANs as their computation cost scales linearly and no additional relation encoders are required. However, the locality nature of GCNs precludes efficient non-local information propagation. To address this issue, we propose the dynamic fusion mechanism, which integrates higher order information for better non-local information aggregation. With the help of this mechanism, our model solely based on graph convolutions is able to outperform competitive structured SANs.

Inspired by Gated Linear Units (GLUs; \citealt{Dauphin2016LanguageMW}), which leverage gating mechanisms \citep{hochreiter1997long} to dynamically control information flows in the convolutional neural networks, we propose dynamic fusion mechanism (DFM) to integrate information from different orders. DFM allows the model to automatically synthesize information from neighbors at varying degrees of hops away. Similar to GLUs, DFM retains non-linear capabilities of the layer while allowing the gradient to propagate through the linear unit without scaling. Based on this non-linear mixture procedure, DFM is able to control the information flows from a range of orders to specific nodes in the AMR graph. Formally, graph convolutions based on DFM are defined as:
\begin{equation}
\small
\begin{aligned}
\vspace{-1mm}
\mathbf{H}_{l+1} = & (1 - \frac{1}{K-1}\sum\limits_{ 1<k<K}\mathbf{G}^{(k)}_l) \odot \phi (\mathbf{A} \mathbf{H}_{l} \mathbf{W}_{l}  + \mathbf{b}_l )  \\
& + \frac{1}{K-1} \sum\limits_{1<k<K}\mathbf{G}^{(k)}_l \odot \phi (\mathbf{A}^k\mathbf{H}_{l}  \mathbf{W}_{l}  + \mathbf{b}_l ).
\vspace{-1mm}
\end{aligned}
\label{hgcn}
\end{equation}
where $\mathbf{G}^{(k)}_l$ is a gating matrix conditioned on the $k$-th order adjacency matrix $\mathbf{A}^{k}$, namely:
\begin{equation}
\vspace{-1mm}
\mathbf{G}^{(k)}_l = (1 - \lambda^{k}) \odot \sigma(\mathbf{A}^k \mathbf{H}_{l}  \mathbf{W}_{l}  + \mathbf{b}_l ),
\vspace{-1mm}
\end{equation}
where $\odot$ denotes elementwise product, $\sigma$ denotes the sigmoid function, $\lambda \in (0,1)$ is a scalar, $K\geq2$ is the highest order used for information aggregation, and $\mathbf{W}_{l}$ denotes trainable weights shared by different $\mathbf{A}^{k}$. Both $\lambda$ and $K$ are hyperparameters.

\paragraph{Computational Overhead}
\label{sec:4.1.2}
In practice, there is no need to calculate or store $\mathbf{A}^{k}$. $\mathbf{A}^{k}\mathbf{H}_{l}$ is computed with right-to-left multiplication. Specifically, if $k$=3, we calculate $\mathbf{A}^{3}\mathbf{H}_{l}$ as $(\mathbf{A}(\mathbf{A}(\mathbf{A}\mathbf{H}_{l})))$. Since we store $\mathbf{A}$ as a sparse matrix with $m$ non-zero entries as vanilla GCNs, an efficient implementation of our layer takes $O(k_{max} \times m \times d)$ computational time, where $k_{max}$ is the highest order used and $d$ is the feature dimension of $\mathbf{H}_{l}$. Under the realistic assumptions of $k_{max} \ll m$ and $d \ll m$, running an $l$-layer model takes $O(lm)$ computational time. This matches the computational complexity of the vanilla GCNs. On the other hand, DFM does not require additional parameters as the weight matrix is shared over various orders.

\paragraph{Deeper LDGCNs}
To further facilitate the non-local information aggregation, we stack several LDGCN layers. In order to stabilize the training, we introduce dense connections \citep{Huang2017DenselyCC, guo2019densely} into the LDGCN model. Mathematically, we define the input of the $l$-th layer $\mathbf{\hat{H}}_{l}$ as the concatenation of all node representations produced in layers $1$, $\cdots$, $l-1$:
\begin{equation}
\vspace{-1mm}
\mathbf{\hat{H}}_{l} = [\mathbf{H}_{0};\mathbf{H}_{1}; ... ;\mathbf{H}_{l-1}]. 
\label{dense}
\vspace{-1mm}
\end{equation}
~~Accordingly, $\mathbf{H}_{l}$ in Eq.~\ref{hgcn} is replaced by $\mathbf{\hat{H}}_{l}$. 
\begin{equation}
\small
\begin{aligned}
\mathbf{H}_{l+1} = & (1 - \frac{1}{K-1}\sum\limits_{ 1<k<K}\mathbf{G}^{(k)}_l) \odot \phi (\mathbf{A} \mathbf{\hat{H}}_{l} \mathbf{W}_{l}  + \mathbf{b}_l )  \\
& + \frac{1}{K-1} \sum\limits_{1<k<K}\mathbf{G}^{(k)}_l \odot \phi (\mathbf{A}^k \mathbf{\hat{H}}_{l}  \mathbf{W}_{l}  + \mathbf{b}_l ).
\end{aligned}
\end{equation}
where $\textbf{W}_{l} \in \mathbb{R}^{d_{l} \times d}$ and $d_{l}$=$d \times (l-1)$. The model size scales linearly as we increase the depth of the network.



\begin{figure}
	\centering
	\includegraphics[scale=0.35]{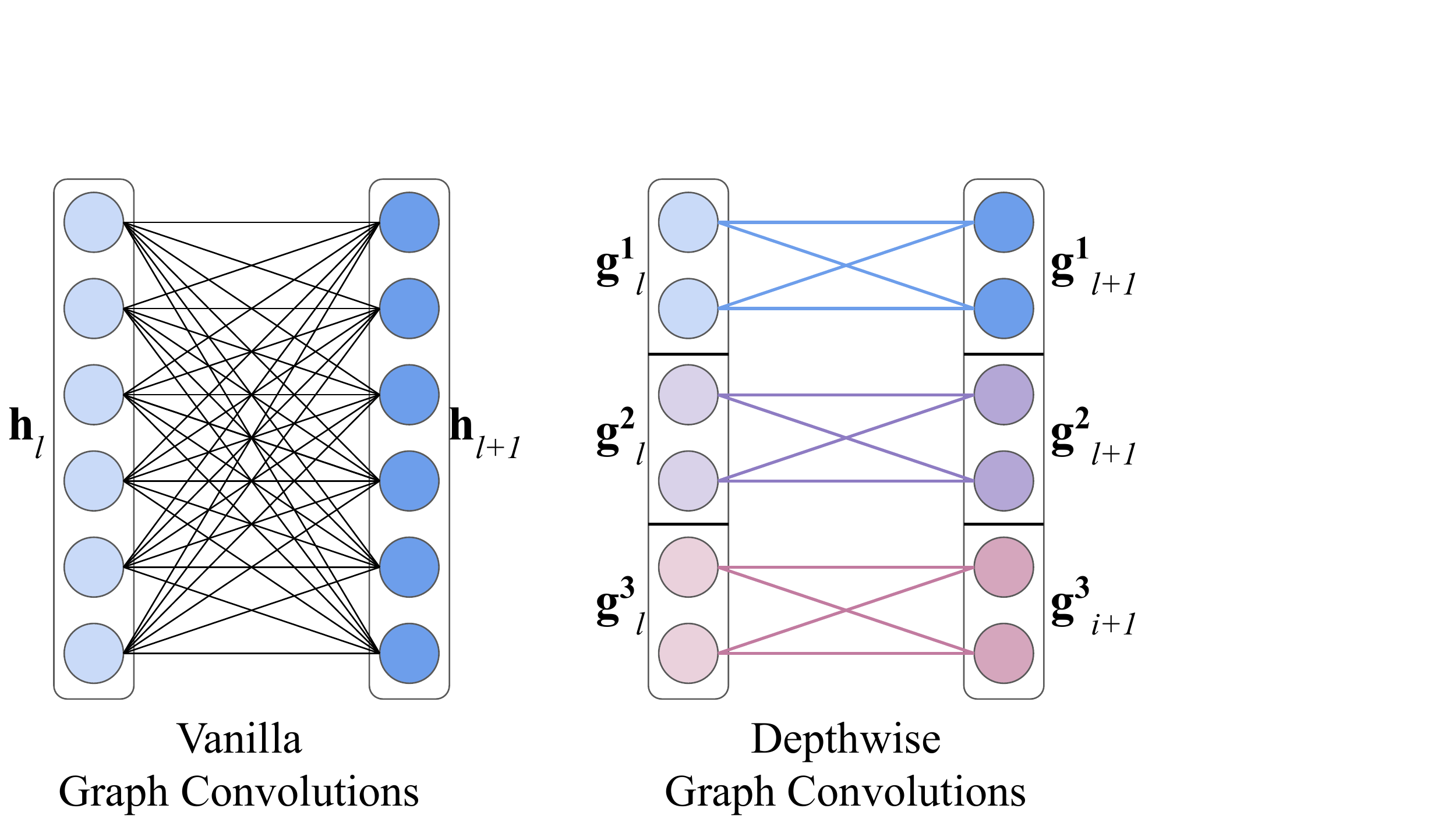}
	\vspace{-1mm}
	\caption{Comparison between vanilla graph convolutions and depthwise graph convolutions. The input and output representation of the $l$-th layer $\mathbf{h}_{l}$ and $\mathbf{h}_{l+1}$ are partitioned into $N$=$3$ disjoint groups.}
	\vspace{-5mm}
	\label{fig:depth}
\end{figure}

\section{Parameter Saving Strategies}
\label{sec:3}
Although we are able to train a very deep LDGCN model, the LDGCN model size increases sharply as we stack more layers, resulting in large model complexity. To maintain a better balance between parameter efficiency and model capacity, we develop two novel parameter saving strategies. We first reduce partial parameters in each layer based on group graph convolutions. Then we further share parameters across all layers based on weight tied convolutions. These strategies allow the LDGCN model to reduce memory usage and model complexity.

\subsection{Group Graph Convolutions}
\label{sec:3.1}

Group convolutions have been used to build efficient networks for various computer vision tasks as they can better integrate feature maps \citep{Xie2017AggregatedRT,Li2019SelectiveKN} and have lower computational costs \citep{Howard2017MobileNetsEC, Zhang2017ShuffleNetAE} compared to vanilla convolutions. In order to reduce the model complexity in the deep LDGCN model, we extend group convolutions to GCNs by introducing group convolutions along two directions: depthwise and layerwise.

\paragraph{Depthwise Graph Convolutions: }
\label{sec:3.1.1}
As discussed in Section \ref{sec:2}, graph convolutions operate on the features of $n$ nodes $\mathbf{H} \in \mathbb{R}^{n \times d}$. For simplicity, we assume $n$=1, the input and output representation of the $l$-th layer are $\mathbf{h}_{l} \in \mathbb{R}^{d_l}$ and $\mathbf{h}_{l+1} \in \mathbb{R}^{d_{l+1}}$, respectively. As shown in Figure \ref{fig:depth}, the size of the weight matrix $\mathbf{W}_{l}$ in a vanilla graph convolutions is $d_{l} \times d_{l+1}$. In depthwise graph convolutions, $\mathbf{h}_{l}$ is partitioned into $N$ mutually exclusive groups $\{\mathbf{g}^{1}_{l}$,...,$\mathbf{g}^{N}_{l}\}$. The weight $\mathbf{W}_{l}$ of each layer is also partitioned into $N$ mutually exclusive groups $\mathbf{
	W}^{1}_{l}$,...,$\mathbf{W}^{N}_{l}$. The dimension of each weight is $\frac{d_{l}}{N} \times \frac{d_{l+1}}{N}$. Finally, we obtain the output representation $\mathbf{h}_{l+1}$ by concatenating $N$ groups of outputs [$\mathbf{g}^{1}_{l+1}$;...;$\mathbf{g}^{N}_{l+1}$]. Now the parameters of each layer can be reduced by a factor of $N$, to $\frac{d_{l} \times d_{l+1}}{N}$.




\paragraph{Layerwise Graph Convolutions: }
\label{sec:3.1.2}

\begin{figure}
	\centering
	\includegraphics[scale=0.35]{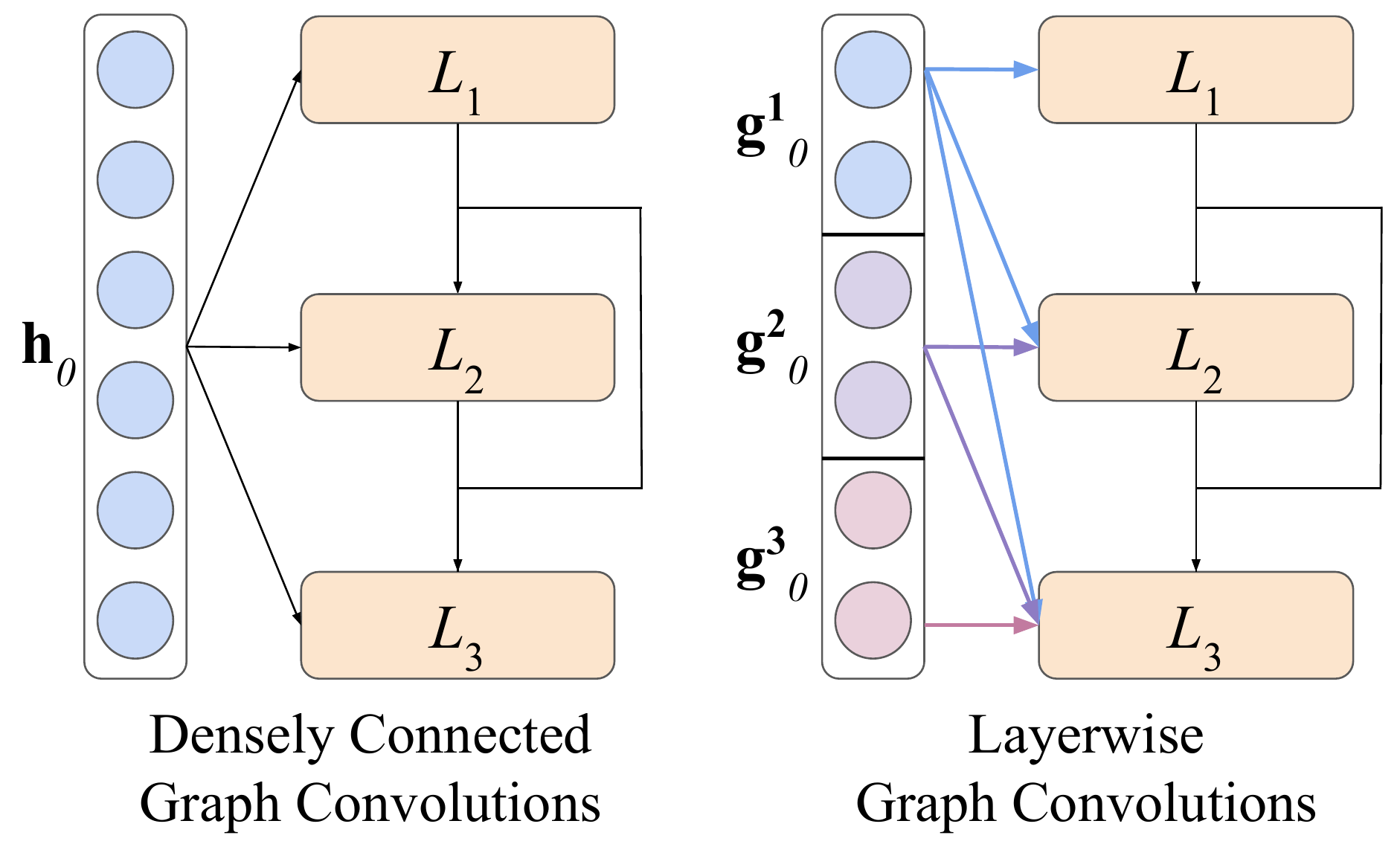}
	\vspace{-2mm}
	\caption{Comparison between vanilla graph convolutions and layerwise graph convolutions. The input representation $\mathbf{h}_{0}$ is partitioned into $M$=$3$ disjoint groups.}
	\vspace{-5mm}
	\label{fig:layerwise}
\end{figure}

These group convolutions are built based on densely connected graph convolutions \citep{guo2019densely}. As shown in Figure \ref{fig:layerwise}, each layer takes the concatenation of outputs from all preceding layers as its input. For example, layer $L_{2}$ takes the concatenation of [$\mathbf{h}_{0};\mathbf{h}_{1}$] as its input. \citet{guo2019densely} further adopt a dimension shrinkage strategy. Assume $\mathbf{h}_{0} \in \mathbb{R}^{d}$ and that the network has $L$ layers. The dimension of output for each layer is set to $\frac{d}{L}$. Finally, we concatenate the output of $L$ layers [$\mathbf{h}_{1};...;\mathbf{h}_{L}$] to form the final representation $h_{final} \in \mathbb{R}^{d}$. Therefore, the size of weight matrix for $l$-th layer is $(d+\frac{d\times (l-1)}{L} ) \times \frac{d}{L}$.

Notice that main computation cost originates in the computation of $\mathbf{h}_{0}$ as it has a large dimension and it is concatenated to the input of each layer. In layerwise graph convolutions however, we improve the parameter efficiency by dividing the input representation $\mathbf{h}_{0}$ into $M$ groups $\{\mathbf{g}^1_{0}$,...$\mathbf{g}^M_{0}\}$, where $M$ equals the total number of layers $L$. The first group $\mathbf{g}^1_{0}$ is fed to all $L$ layers, and the second group $\mathbf{g}^2_{0}$ is fed to ($L$-1) layers, so on and so forth. Accordingly, the size of weight matrix for the $l$-th layer is $(\frac{d\times (2l-1)}{L} ) \times \frac{d}{L}$.



Formally, we partition the input representations of $n$ concept $\mathbf{H}_{0} \in \mathbb{R}^{n \times d}$ to the first layer into $M$ groups $\{ \mathbf{G}^{1}_{0},..., \mathbf{G}^{M}_{0}\}$, where the size of each group is $n \times \frac{d}{M}$. Accordingly, we modify the input of the $l$-th layer $\mathbf{\hat{H}}_{l}$ in Eq.~\ref{dense} as:
\begin{equation}
\mathbf{\hat{H}}_{l} = [\mathbf{G}^{1}_{0};...; \mathbf{G}^{M}_{0};\mathbf{H}_{1}; ... ;\mathbf{H}_{l-1}]
\end{equation}

In practice, we combine these two convolutions together to further reduce the model size. For example, assume the size of the input is $d$=360 and the number of layers is $L$=6.  The size of the weight matrix for the first layer ($l$=1) is $(d+\frac{d\times (l-1)}{L} ) \times \frac{d}{L}$=$360 \times 60$. Assume we set $N$=3 for depthwise graph convolutions and $M$=6 for layerwise graph convolutions. We first use layerwise graph convolutions by dividing the input into 6 groups, where each one has the size $\frac{d}{M}$=60. Then we feed the first group to the first layer. Next we use depthwise graph convolutions to further split the input into 3 groups. We now have 3 weight matrices for the first layer, each one with the size $\frac{d\times (2l-1)}{M} \times \frac{d}{M \times N}$= $20 \times 20$. With the increase of the feature dimension $d$ and the number of layer $L$, more prominent parameter efficiency can be observed.



\subsection{Weight Tied Convolutions}
\label{sec:3.2}
We further adopt a more aggressive strategy where parameters are shared across all layers. This further significantly reduces the size of the model. Theoretically, weight tied networks can be unrolled to any depth, typically with improved feature abstractions as depth increases \citep{Bai2019DeepEM}. Recently, weight tied SANs were explored to regularize the training and help with generalization \citep{Dehghani2019UniversalT, Lan2020ALBERTAL}. Mathematically, Eq.~\ref{eq:graphconv} can be rewritten as:
\begin{equation}
\begin{aligned}
\vspace{-1mm}
\mathbf{H}_{l+1} = \phi (\mathbf{A} \mathbf{\hat{H}}_{l} \mathbf{W} +  \mathbf{b}),
\label{eq:weighttied}
\vspace{-1mm}
\end{aligned}
\end{equation}
where $\mathbf{W}$ and $\mathbf{b}$ are shared parameters for all convolutional layers. To stabilize training, a gating mechanism was introduced to graph neural networks in order to build graph recurrent networks \citep{Li2015GatedGS,Song2018AGM}, where parameters are shared across states (time steps). However, the graph convolutional structure is very deep (e.g., 36 layers). Instead, we adopt a jumping connection \citep{Xu2018RepresentationLO}, which forms the final representation $\mathbf{H}_{final}$ based on the output of all layers. This connection mechanism can be considered deep supervision \citep{Lee2015DeeplySupervisedN, Bai2019TrellisNF} for training deep convolutional neural networks. Formally, the $\mathbf{H}_{final}$ of LDGCNs which have $L$ layer is obtained by:
$\mathbf{H}_{final} =  \mathcal{F}(\mathbf{\hat{H}}_{L},...,\mathbf{\hat{H}}_{1})$,
where $\mathcal{F}$ is a linear transformation.

\section{Experiments}
\label{sec:5}

\subsection{Setup}
\label{ssec:5.1}
We evaluate our model on the  LDC2015E86 (AMR1.0), LDC2017T10 (AMR2.0) and LDC2020T02 (AMR3.0) datasets, which have 16,833, 36,521 and 55,635 instances for training, respectively. Both AMR1.0 and AMR2.0 have 1,368 instances for development, and 1,371 instances for testing. AMR3.0 has 1,722 instances for development and 1,898 instances for testing. Following \citet{Zhu2019ModelingGS}, we use byte pair encodings \cite{Sennrich2016NeuralMT} to deal with rare words.







Following \citet{guo2019densely}, we stack 4 LDGCN blocks as the encoder of our model.  Each block consists of two sub-blocks where the bottom one contains 6 layers and the top one contains 3 layers. The hidden dimension of LDGCN model is 480. Other model hyperparameters are set as $\lambda$=0.7, $K$=2 for dynamic fusion mechanism, $N$=2 for depthwise graph convolutions and $M$=6 and 3 for layerwise graph convolutions for the bottom and top sub-blocks, respectively. For the decoder, we employ the same attention-based LSTM as in previous work \citep{Beck2018GraphtoSequenceLU, guo2019densely, Damonte2019StructuralNE}. Following \citet{Wang2020AMRToTextGW}, we use a transformer as the decoder for large-scale evaluation. For fair comparisons, we use the same optimization and regularization strategies as in \citet{guo2019densely}. All hyperparameters are tuned on the development set\footnote{Hyperparameter search; all hyperparameters are attached in the supplementary material.}.  



\begin{table*}[!th]
	\small
	\centering
	\setlength{\tabcolsep}{2.5pt}
	\begin{tabular}{lccccccccc} 
		\toprule
		\multirow{2}{*}{{Model}} & \multirow{2}{*}{{Type}} & \multicolumn{4}{c}{{AMR2015}} & \multicolumn{4}{c}{{AMR2017}} \\
		\cmidrule(l{5pt}r{5pt}){3-6} \cmidrule(l{5pt}r{5pt}){7-10}
		& & B & C & M & \#P 
		& B & C & M & \#P   \\
		\midrule
		Seq2Seq~\cite{Cao2018FactorisingAG}     &Single  & 23.5 & -  &- & -  & 26.8  &- & - & -\\
		GraphLSTM~\cite{Song2018AGM}              &Single & 23.3 & -  &- & - & 24.9  &- & - & -\\
		GGNNs~\cite{Beck2018GraphtoSequenceLU} &Single & - & -  &- &- & 23.3  &50.4 & - & 28.3M\\
		GCNLSTM~\cite{Damonte2019StructuralNE} &Single & 24.4 & -  &23.6 & - & 24.5  &- &24.1 & {\color{white}0}30.8M$^{\ddagger}$ \\
		
		DCGCN~\cite{guo2019densely}      &Single  & 25.7 & {\color{white}0} 54.5$^{\ddagger}$  &{\color{white}0}31.5$^{\ddagger}$  & {\color{white}0}18.6M$^{\ddagger}$ & 27.6  &57.3 & 34.0 & 19.1M\\
		
		DualGraph~\cite{Ribeiro2019EnhancingAG}     &Single  &24.3 & {\color{white}0}53.8$^{\ddagger}$  &30.5 & {\color{white}0}60.3M$^{\ddagger}$ & 27.9  & {\color{white}0}58.0$^{\ddagger}$  &33.2  & {\color{white}0}61.7M$^{\ddagger}$ \\
		
		\midrule
		Seq2Seq~\cite{Konstas2017NeuralAS}      &Ensemble  & - & -  &- & - & 26.6  &52.5 & - & 142M\\
		GGNNs~\cite{Beck2018GraphtoSequenceLU} &Ensemble & - & -  &- & - & 27.5  &53.5 & - & 141M\\
		DCGCN~\cite{guo2019densely}      &Ensemble  & - & -  &- & - & 30.4  &59.6 & - & 92.5M\\
		
		\midrule
		Transformer~\cite{Zhu2019ModelingGS}  &Single &  25.5 & 59.9  & 33.1 & 49.1M  & 27.4 & 61.9 & 34.6 & - \\
		GT\_Dual~\cite{Wang2020AMRToTextGW}  &Single &  25.9 & -  &- & 19.9M  & 29.3  &59.0 & - & 19.9M \\
		GT\_GRU~\cite{Cai2019GraphTF}       &Single & 27.4 & 56.4  &32.9 & 30.8M  & 29.8  &59.4 & 35.1 & 32.2M\\
		GT\_SAN~\cite{Zhu2019ModelingGS}        &Single & 29.7 & {\color{white}0}60.7$^{\ddagger}$  &35.5 & 49.3M  & 31.8  &{\color{white}0}61.8$^{\ddagger}$ & 36.4 & {\color{white}0}54.0M$^{\ddagger}$\\
		\midrule
		LDGCN\_WT                         &Single   & {28.6} &{58.5}  &{33.1} & \textbf{10.6M} & {31.9}  &{61.2} & {36.3} & \textbf{11.8M}\\
		LDGCN\_GC                        &Single  & \textbf{30.8} &\textbf{61.8}  &\textbf{36.4} & {12.9M} & \textbf{33.6}  &\textbf{63.2} & \textbf{37.5} & {13.6M}\\
		
		\bottomrule
	\end{tabular}
	\vspace{-2mm}
	\caption{Main results on AMR-to-text generation. B, C, M and \#P  denote BLEU, CHRF++, METEOR and the model size in terms of parameters, respectively. Results with $\ddagger$ are obtained from the authors. We also conduct the statistical significance tests by following \citep{Zhu2019ModelingGS}. All our proposed systems are significant over the baseline at $p<0.01$, tested by bootstrap resampling \citep{Koehn2004StatisticalST}. }
	\vspace{-3mm}
	\label{tab:amr}
\end{table*}


\begin{table}[!t]
	\small
	\centering
	\setlength{\tabcolsep}{3pt}
	\begin{tabular}{lccc}
		\toprule
		Model & \#P &External &B \\
		\midrule
		Seq2Seq~\citep{Konstas2017NeuralAS} & - & ~~~2M & 32.3 \\
		Seq2Seq~\citep{Konstas2017NeuralAS} & - & ~~20M & 33.8 \\
		GraphLSTM~\citep{Song2018AGM} & - & ~~~2M & 33.6 \\
		Transformer~\cite{Wang2020AMRToTextGW} & - & ~~~2M & 35.1 \\
		GT\_Dual~\cite{Wang2020AMRToTextGW} & 78.4M & ~~~2M & 36.4 \\ 
		\midrule
		LDGCN\_GC & 23.2M & 0.5M & 36.0  \\
		LDGCN\_WT & \textbf{20.8M} & \textbf{0.5M} & \textbf{36.8} \\
		\bottomrule
	\end{tabular}
	\caption{Results on AMR1.0 with external training data. $^{\ddagger}$ denotes the ensemble model.}
	\vspace{-6mm}
	\label{tab:external}
\end{table}

For evaluation, we report BLEU scores~\cite{Papineni2002BleuAM}, CHRF++~\cite{Popovic2017chrFWH} scores and METEOR scores~\cite{Denkowski2014MeteorUL} with additional human evaluation results.



\subsection{Main Results}
\label{sec:5.2}
We consider two kinds of baseline models:  1)  models based on Recurrent Neural Networks ~\cite{Konstas2017NeuralAS,Cao2018FactorisingAG} and Graph Neural Networks (GNNs)~\cite{Song2018AGM, Beck2018GraphtoSequenceLU, Damonte2019StructuralNE,guo2019densely,Ribeiro2019EnhancingAG}. These models use an attention-based LSTM decoder. 2) models based on SANs~\cite{Zhu2019ModelingGS} and structured SANs~\cite{Cai2019GraphTF,Zhu2019ModelingGS,Wang2020AMRToTextGW}. Specifically, \citet{Zhu2019ModelingGS} leverage additional SANs to incorporate the relational encoding whereas \citet{Cai2019GraphTF} use GRUs.
Additional results of ensemble models are also included. The results are reported in Table \ref{tab:amr}. Our model has two variants based on different parameter saving strategies, including LDGCN\_WT (weight tied) and LDGCN\_GC (group convolutions), and both of them use the dynamic fusion mechanism (DFM).

\paragraph{LDGCN v.s. Structured SANs.}  Compared to state-of-the-art structured SANs (GT\_SAN), the performance of LDGCN\_GC is 1.1 and 1.8 BLEU points higher on AMR1.0 and AMR2.0, respectively. Moreover, LDGCN\_GC  requires only about a quarter of the number of parameters (12.9M vs 49.0M, and 13.6M vs 54.0M). Our more lightweight variant LDGCN\_WT achieves better BLEU scores than GT\_SAN on AMR2.0 while using only 1/5 of their model parameters. However, LDGCN\_WT obtains lower scores on AMR1.0 than GT\_SAN. We hypothesize that weight tied convolutions require more data to train as we observe severe oscillations when training the model on the small AMR1.0 dataset. The oscillation is reduced when we train it on the larger AMR2.0 dataset and the semi-supervised dataset. 

\begin{table}[!t]
	\small
	\centering
	\setlength{\tabcolsep}{3pt}
	\begin{tabular}{lcccc}
		\toprule
		\bf Model  &B &C &M & \#P  \\
		\midrule
		GGNNs \citep{Beck2018GraphtoSequenceLU}  &26.7$^\dagger$ &57.2$^\dagger$ &33.1$^\dagger$ & 30.9M$^\dagger$ \\
		DCGCN \citep{guo2019densely}  & 29.8$^\ddagger$ &59.9$^\ddagger$ &35.6$^\ddagger$ &22.2M$^\ddagger$ \\
		\midrule
		LDGCN\_WT   &33.0 &62.6 &36.5  & \textbf{11.5M}\\
		LDGCN\_GC   & \textbf{34.3} & \textbf{63.7} & \textbf{38.2} &14.3M  \\
		\bottomrule
	\end{tabular}
	\caption{Results on the AMR3.0. B, C, M and \#P  denote BLEU, CHRF++, METEOR and the model size in terms of parameters, respectively. The results with $\dagger$ are based on open implementations, while the results with $\ddagger$ are obtained from the authors.}
	\vspace{-4mm}
	\label{tab:ws}
\end{table}


\paragraph{LDGCN v.s. Other GNNs.} Both LDGCN models significantly outperform GNN-based models. For example, LDGCN\_GC surpasses DCGCN by 5.1 points on AMR1.0 and surpasses DualGraph by 5.7 points on AMR2.0. Moreover, the single LDGCN model achieves consistently better results than previous ensemble GNN-based models in BLEU, CHRF++ and METEOR scores. In particular, on AMR2.0, LDGCN\_WT obtains 1.5 BLEU points higher than the DCGCN ensemble model, while requiring only about 1/8 of the number of parameters. We also evaluate our model on the latest AMR3.0 dataset. Results are shown in Table~\ref{tab:ws}. LDGCN\_WT and LDGCN\_GC consistently outperform GNN-based models including DCGCN and GGNNs on this larger dataset. These results suggest that LDGCN can learn better representation more efficiently. 

\paragraph{Large-scale Evaluation.} We further evaluate LDGCNs on a large-scale dataset. Following \citet{Wang2020AMRToTextGW}, we first use the additional data to pretrain the model, then finetune it on the gold data. Evaluation results are reported in Table.\ref{tab:external}. 
Using 0.5M data, LDGCN\_WT outperforms all models including structured SANs with 2M additional data. These results show that our model is more effective in terms of using a larger dataset. Interestingly, LDGCN\_WT consistently outperforms LDGCN\_GC under this setting. Unlike training the model on AMR1.0, training LDGCN\_WT on the large-scale dataset has fewer oscillations, which confirms our hypothesis that sufficient data acts as a regularizer to stabilize the training process of weight tied models.

\subsection{Development Experiments} 
We conduct an ablation study to demonstrate how dynamic fusion mechanism and parameter saving strategies are beneficial to the lightweight model with better performance based on development of experimental results on AMR1.0. Results are shown in Table~\ref{tab:ablation}. DeepGCN is the model with dense connections \citep{Huang2017DenselyCC,guo2019densely}. DeepGCN+GC+DF and DeepGCN+WT+DF are essentially LDGCN\_GC and LDGCN\_WT models in Section \ref{sec:5.2}, respectively.


\begin{table}[!t]
	\small
	\centering
	\setlength{\tabcolsep}{3pt}
	\begin{tabular}{lcc}
		\toprule
		Model &\#Parameters  &BLEU \\
		\midrule
		DeepGCN               &19.9M  &29.3\\
		DeepGCN+DF         &19.9M  &30.4 \\
		DeepGCN+GC         &12.9M  &29.0 \\
		DeepGCN+GC+DF (LDGCN\_GC)    &12.9M  &30.3 \\
		DeepGCN+WT         &10.6M  &27.4 \\
		DeepGCN+WT+DF (LDGCN\_WT)   &10.6M  &28.3\\
		\bottomrule
	\end{tabular}
	\caption{Comparisons between baselines. +DF denotes dynamic fusion mechanism. +WT and +GC refer to weight tied and group convolutions, respectively.}
	\label{tab:ablation}
\end{table}

\begin{table}[!t]
	\small
	\centering
	\setlength{\tabcolsep}{3pt}
	\begin{tabular}{lcc}
		\toprule
		Model  & Inference Speed \\
		\midrule
		Transformer                & 1.00x\\
		DeepGCN          & 1.21x \\
		LDGCN\_WT           & 1.22x \\
		LDGCN\_GC     & 1.17x \\
		\bottomrule
	\end{tabular}
	\caption{Speed comparisons between baselines. For inference speed, the higher the better. Implementations are based on MXNet \citep{Chen2015MXNetAF} and the Sockeye neural machine translation toolkit \citep{hieber2017sockeye}. Results on speed are based on beam size 10, batch size 30 on an NVIDIA RTX 1080 GPU.}
	\vspace{-4mm}
	\label{tab:speed}
\end{table}

\paragraph{Dynamic Fusion Mechanism.} The performance of DeepGCN+DF is 1.1 BLEU points higher than DeepGCN, which demonstrates that our dynamic fusion mechanism is beneficial for graph encoding when applied alone. Adding the group graph convolutions strategies gives a BLEU score of 30.3, which is only 0.1 points lower than DeepGCN+DF. This result shows that the representation learning ability of the dynamic fusion mechanism is robust against parameter sharing and reduction. We also observe that the mechanism helps to alleviate oscillation when training the weight tied model. DeepGCN+WT+DF achieves better results than DeepGCN+WT, which is hard to converge when training it on the small AMR1.0 dataset.

\paragraph{Parameter Saving Strategy.} Table~\ref{tab:ablation} demonstrates that although the performance of DeepGCN+GC is only 0.3 BLEU points lower than that of DeepGCN, DeepGCN+GC only requires $65\%$ of the number of parameters of DeepGCN. Furthermore, by introducing the dynamic fusion mechanism, the performance of DeepGCN+GC is improved greatly and is in fact on par with DeepGCN. Also, DeepGCN+GC+DF does not rely on any kind of self-attention layers, hence, its number of parameters is much smaller than that of graph transformers, i.e., DeepGCN+GC+DF only needs $1/4$ to $1/3$ the number of parameters of graph transformers, as shown in Table~\ref{tab:amr}. On the other hand, DeepGCN+WT is more efficient than DeepGCN+GC. As shown in Table \ref{tab:external}, with an increase in training data, more prominent parameter efficiency can be observed. 

\paragraph{Time Cost Analysis.} As shown in the Table \ref{tab:speed}, all three GCN-based models outperform the SAN-based model in terms of speed because the computation of attention weights scales quadratically while convolutions scale linearly with respect to the input graph size. LDGCN\_GC is slightly slower than the other two models, since it requires additional tensor split operations. We believe that state-of-the-art structured SANs are also strictly slower than vanilla SANs, as they require additional neural components, such as GRUs, to encode  structural information in the AMR graph. In summary, our model not only has better parameter efficiency, but also lower time costs.

\subsection{Human Evaluation}
We further assess the quality of the generated sentences with human evaluation. Following \citet{Ribeiro2019EnhancingAG}, two evaluation criteria are used: (i) meaning similarity:  how close in meaning the generated text is to the gold sentence; (ii) readability: how well the generated sentence reads. We randomly select 100 sentences generated by 4 models. 30 human subjects rate the sentences on a 0-100 rating scale. The evaluation is conducted separately and subjects were first given brief instructions explaining the criteria of assessment. For each sentence, we collect scores from 5 subjects and average them. Models are ranked according to the mean of sentence-level scores. Also, we apply a quality control step filtering subjects who do not score some faked and known sentences properly.

As shown in Table \ref{tab:human}, LDGCN\_GC has better human rankings in terms of both meaning similarity and readability than the state-of-the art GNN-based (DualGraph) and SAN-based model (GT\_SAN). DeepGCN without dynamic fusion mechanism obtains lower scores than GT\_SAN, which further confirms that synthesizing higher order information helps in learning better graph representations.

\begin{table}[!t]
	\small
	\centering
	\setlength{\tabcolsep}{3pt}
	\begin{tabular}{lcc}
		\toprule
		Model & Similarity  &Readability \\
		\midrule
		DualGraph~\citep{Ribeiro2019EnhancingAG}            & 65.07  & 68.78 \\
		GT\_SAN~\citep{Zhu2019ModelingGS}      & 69.63  & 72.23 \\
		DeepGCN      & 68.91  & 71.45 \\
		LDGCN\_GC      & \textbf{71.92}  &\textbf{74.16} \\
		\bottomrule
	\end{tabular}
	\caption{Human evaluation. We also perform significance tests by following \citep{Ribeiro2019EnhancingAG}. Results are statistically significant with $p<0.05$.}
	\vspace{-5mm}
	\label{tab:human}
\end{table}

\begin{figure}[!t]
	\centering
	\includegraphics[scale=0.4]{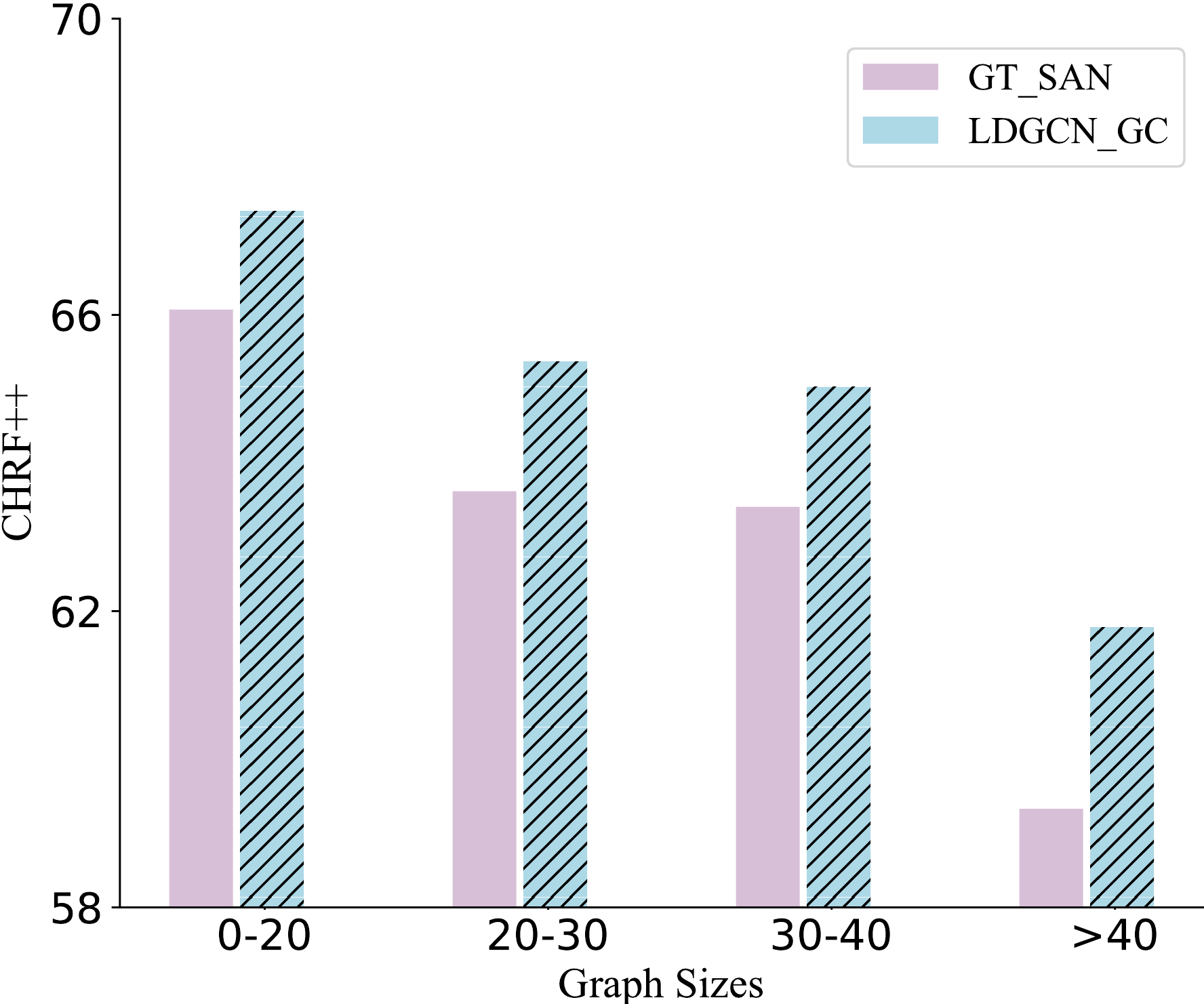}
	\vspace{-2mm}
	\caption{Performance against graph sizes.}
	\label{fig:graph_size}
\end{figure}

\begin{figure}[!t]
	\centering
	\includegraphics[scale=0.4]{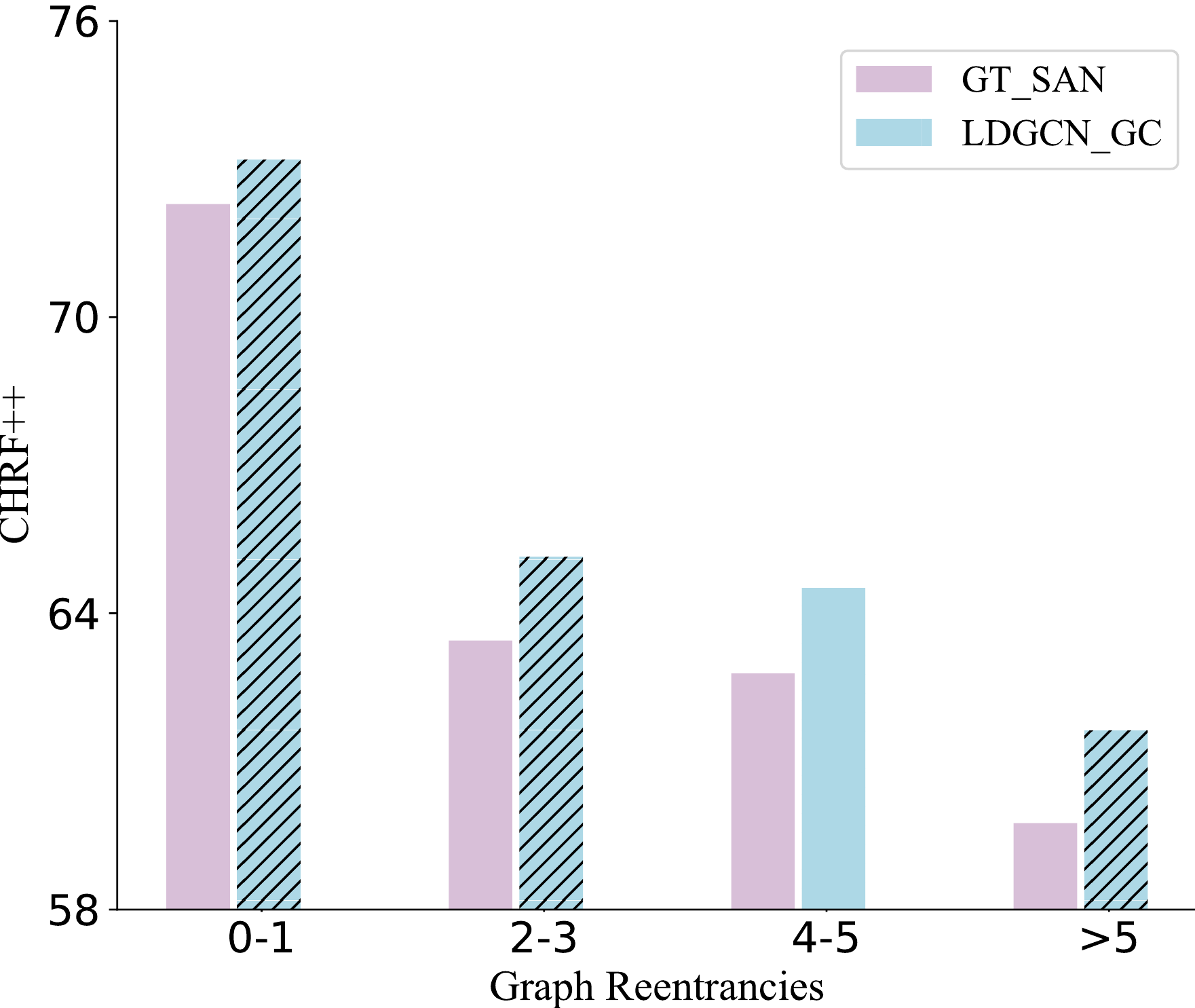}
	\vspace{-2mm}
	\caption{Performance against graph re-entrancies.}
	\label{fig:reen}
	\vspace{-3mm}
\end{figure}

\subsection{Additional Analysis}

To further reveal the source of performance gains, we perform additional analysis based on the characteristics of AMR graphs, i.e., graph size and graph reentrancy ~\cite{Damonte2019StructuralNE,Damonte20}. All experiments are conducted on the AMR2.0 test set and  CHRF++ scores are reported.

\paragraph{Graph Size.} As shown in Figure~\ref{fig:graph_size}, the size of AMR graphs is partitioned into four categories (($0$, $20$], ($20$, $30$], ($30$, $40$], $>40$),  Overall, LDGCN\_GC outperforms the best-reported GT\_SAN model across all graph sizes, and the performance gap becomes more profound with the increase of graph sizes. Although both models have sharp performance degradation for extremely large graphs ($>40$), the performance of LDGCN\_GC is more stable. Such a result suggests that our model can better deal with large graphs with more complicated structures. 

\paragraph{Graph re-entrancies.} Reentrancies describe the co-references and control structures in AMR graphs. A graph is considered more complex if it contains more re-entrancies. In Figure~\ref{fig:reen}, we show how the LDGCN\_GC and GT\_SAN generalize to different scales of reentrancies. Again, LDGCN\_GC consistently outperforms GT\_SAN and the performance gap becomes noticeably wider when the number of re-entrancies increases. These results suggest that our model can better model the complex dependencies in AMR graphs. 

\begin{table}[!t]
	\small
	\centering
	\setlength{\tabcolsep}{3pt}
	\begin{tabular}{lccc}
		\toprule
		(m / multi-sentence \\
		{\color{white}0}{\color{white}0}{\color{white}0}   
		:snt1 (t / trust-01  \\ 
		{\color{white}0}{\color{white}0}{\color{white}0}{\color{white}0}{\color{white}0}{\color{white}0}   
		:ARG2 (i / i))  \\
		{\color{white}0}{\color{white}0}{\color{white}0}
		:snt2 (g / good-02   \\
		{\color{white}0}{\color{white}0}{\color{white}0}{\color{white}0}{\color{white}0}{\color{white}0}
		:ARG1 (g2 / \textbf{get-01} \\
		{\color{white}0}{\color{white}0}{\color{white}0}{\color{white}0}{\color{white}0}{\color{white}0}{\color{white}0}{\color{white}0}{\color{white}0}
		:ARG1 (t2 / thing   \\
		{\color{white}0}{\color{white}0}{\color{white}0}{\color{white}0}{\color{white}0}{\color{white}0}{\color{white}0}{\color{white}0}{\color{white}0}{\color{white}0}{\color{white}0}{\color{white}0}
		:mod (t3 / this))    \\
		{\color{white}0}{\color{white}0}{\color{white}0}{\color{white}0}{\color{white}0}{\color{white}0}{\color{white}0}{\color{white}0}{\color{white}0}   
		:time~e.10,12 (e / \textbf{early}  \\
		{\color{white}0}{\color{white}0}{\color{white}0}{\color{white}0}{\color{white}0}{\color{white}0}{\color{white}0}{\color{white}0}{\color{white}0}{\color{white}0}{\color{white}0}{\color{white}0}
		:degree (m2 / most)  \\
		{\color{white}0}{\color{white}0}{\color{white}0}{\color{white}0}{\color{white}0}{\color{white}0}{\color{white}0}  {\color{white}0}{\color{white}0}{\color{white}0}{\color{white}0}{\color{white}0}:compared-to (p / possible-01  \\
		{\color{white}0}{\color{white}0}{\color{white}0}{\color{white}0}{\color{white}0}{\color{white}0}{\color{white}0}{\color{white}0}{\color{white}0}{\color{white}0}{\color{white}0}{\color{white}0}{\color{white}0}{\color{white}0}{\color{white}0}
		:ARG1 g2))   \\
		{\color{white}0}{\color{white}0}{\color{white}0}{\color{white}0}{\color{white}0}{\color{white}0}{\color{white}0}{\color{white}0}{\color{white}0}
		:ARG1-of (i2 / instead-of-91  \\
		{\color{white}0}{\color{white}0}{\color{white}0}{\color{white}0}{\color{white}0}{\color{white}0}{\color{white}0}{\color{white}0}{\color{white}0}{\color{white}0}{\color{white}0}{\color{white}0}
		:ARG2 (l / let-01  \\
		{\color{white}0}{\color{white}0}{\color{white}0}{\color{white}0}{\color{white}0}{\color{white}0}{\color{white}0}{\color{white}0}{\color{white}0}{\color{white}0}{\color{white}0}{\color{white}0}{\color{white}0}{\color{white}0}{\color{white}0}
		:ARG1 (w / \textbf{worsen-01}  \\
		{\color{white}0}{\color{white}0}{\color{white}0}{\color{white}0}{\color{white}0}{\color{white}0}{\color{white}0}{\color{white}0}{\color{white}0}{\color{white}0}{\color{white}0}{\color{white}0}{\color{white}0}{\color{white}0}{\color{white}0}{\color{white}0}{\color{white}0}{\color{white}0} 
		:ARG1 t2 \\
		{\color{white}0}{\color{white}0}{\color{white}0}{\color{white}0}{\color{white}0}{\color{white}0}{\color{white}0}{\color{white}0}{\color{white}0}{\color{white}0}{\color{white}0}{\color{white}0}{\color{white}0}{\color{white}0}{\color{white}0}{\color{white}0}{\color{white}0}{\color{white}0}
		:mod (e2 / \textbf{even})))))   \\
		{\color{white}0}{\color{white}0}{\color{white}0}{\color{white}0}{\color{white}0}{\color{white}0}{\color{white}0}
		:degree~e.5 (m3 / more)))  \\
		\midrule
		\textbf{Reference}: trust me , it 's better to get  these things as early \\as  possible rather than let them get even worse .\\
		\midrule
		\textbf{DualGraph}: so to me , this is the best thing to get these \\ things as they can , instead of letting it even worse .\\
		\midrule
		\textbf{DeepGCN}: i trust me , it 's better that these things  get \\ in the  early than letting them even get worse . \\
		\midrule
		\textbf{GT\_SAN}: trust me , this is better to get these \\things , rather than let it even get worse . \\
		\midrule
		\textbf{LDGCN\_GC}: trust me . better to get these things  as early \\as possible , rather than letting them even make worse .\\
		\bottomrule
	\end{tabular}
	\caption{An example of AMR graph and generated sentences by different models.}
	\vspace{-4mm}
	\label{tab:sample}
\end{table}

\paragraph{Case Study.} Table~\ref{tab:sample} shows the generated sentence of an AMR graph from four models together with the gold reference. The phrase ``trust me'' is the beginning of the sentence. DualGraph fails to decode it. On the other hand, GT\_SAN successfully generates the second half of the sentence, i.e., ``rather than let them get even worse'', but it fails to capture the meaning of word ``early'' in its output, which is a critical part. DeepGCN parses both ``early'' and ``get even worse'' in the results. However, the readability of the generated sentence is not satisfactory. Compared to  baselines, LDGCN is able to produce the best result, which has a correct starting phrase and captures the semantic meaning of critical words such as ``early'' and ``get even worse'' while also attaining good readability.

\section{Related Work}
\label{sec:4}
Graph  convolutional  networks \citep{Kipf2016SemiSupervisedCW}  have  been  widely used as the structural encoder in various NLP applications including question answering \citep{de2018question, Lin2019KagNetKG}, semantic parsing \citep{Bogin2019GlobalRO,Bogin2019RepresentingSS} and relation extraction \citep{Guo2019AttentionGG, Guo2020LearningLF}.

Early efforts for AMR-to-text generation mainly include grammar-based models \citep{Flanigan2016GenerationFA, Song2017AMRtotextGW} and sequence-based models \citep{Pourdamghani2016GeneratingEF, Konstas2017NeuralAS, Cao2018FactorisingAG}, discarding crucial structural information when linearising the input AMR graph. To solve that, various GNNs including graph recurrent networks \citep{Song2018AGM, Ribeiro2019EnhancingAG} and graph convolutional networks \citep{Damonte2019StructuralNE, guo2019densely} have been used to encode the AMR structure. Though GNNs are able to operate directly on graphs, the locality nature of them precludes efficient information propagation \citep{abu2018n, AbuElHaija2019MixHopHG, luan2019break}. Larger and deeper models are required to model the complex non-local interactions \citep{Xu2018RepresentationLO,li2019can}. More recently, SAN-based models \citep{Zhu2019ModelingGS,Cai2019GraphTF,Wang2020AMRToTextGW} outperform GNN-based models as they are able to capture global dependencies. Unlike previous models, our local, yet efficient model, based solely on graph convolutions, outperforms competitive structured SANs while using a significantly smaller model.

\section{Conclusion}

In this paper, we propose LDGCNs for AMR-to-text generation. Compared with existing GCNs and SANs, LDGCNs maintain a better balance between parameter efficiency and model capacity. LDGCNs outperform state-of-the-art models on AMR-to-text generation. In future work, we would like to investigate methods to stabilize the training of weight tied models and apply our model on other tasks in Natural Language Generation.

\section*{Acknowledgments}
We would like to thank the anonymous reviewers for their constructive comments. We would also like to thank Zheng Zhao, Chunchuan Lyu, Jiangming Liu, Gavin Peng, Waylon Li and Yiluan Guo for their helpful suggestions.
This research is partially supported by Ministry of Education, Singapore, under its Academic Research Fund
(AcRF) Tier 2 Programme (MOE AcRF Tier 2 Award No: MOE2017-T2-1-156). Any opinions,
findings and conclusions or recommendations expressed in this material are those of the authors and
do not reflect the views of the Ministry of Education, Singapore.

\bibliography{emnlp2020}
\bibliographystyle{acl_natbib}

\end{document}